\def\year{2015}
\documentclass{article}
\usepackage{aaai}
\usepackage{times}
\usepackage{helvet}
\usepackage{courier}
\usepackage{amsmath}
\usepackage{amssymb}

\usepackage{amsthm}

\usepackage{graphicx}
\usepackage{multirow}
\usepackage{alltt}

\newcommand{\eat}[1]{}
\def\st#1{\small\tt #1}
\def\textst#1{\text{\st#1}}

\usepackage{url}

\theoremstyle{definition}
\newtheorem{example}{Example}
\newtheorem{definition}{Definition}

\title{Ontology Matching with Knowledge Rules}

\author{
Shangpu Jiang \and Daniel Lowd \and Dejing Dou \\
Computer and Information Science \\
University of Oregon, USA\\
{\it\{shangpu,lowd,dou\}@cs.uoregon.edu}
}

\nocopyright

\begin{document}

\maketitle

\begin{abstract}

Ontology matching is the process of automatically determining the
semantic equivalences between the concepts of two ontologies. Most
ontology matching algorithms are based on two types of strategies:
terminology-based strategies, which align concepts based on their names
or descriptions, and structure-based strategies, which exploit concept
hierarchies to find the alignment. In many domains, there is additional
information about the relationships of concepts represented in various
ways, such as Bayesian networks, decision trees, and association rules.
We propose to use the similarities between these relationships to find
more accurate alignments. We accomplish this by defining soft
constraints that prefer alignments where corresponding concepts have the
same local relationships encoded as {\em knowledge rules}. We use a
probabilistic framework to integrate this new {\em knowledge-based}
strategy with standard terminology-based and structure-based strategies.
Furthermore, our method is particularly effective in identifying
correspondences between
%nominal, numerical attributes and 
complex concepts. Our method achieves substantially better F-score than
the previous state-of-the-art on three ontology matching domains. 

\end{abstract}

%\keywords{ontology matching; knowledge rules; Markov logic; probabilistic
%inference}

\section{Introduction}

Ontology matching is the process of aligning two semantically related
ontologies.  Traditionally, this task is performed by human experts.
Since the task is tedious and error prone, especially in larger
ontologies, there has been substantial work on developing automated or
semi-automated ontology matching systems \cite{Shvaiko11}. While some
automated matching systems make use of data instances, in this paper we
focus on the {\em schema-level} ontology matching task, in which no data
instances are used.

Previous schema-level ontology matching systems mainly use two classes
of strategies. {\em Terminology-based} strategies discover corresponding
concepts with similar names or descriptions. {\em Structure-based}
strategies discover corresponding groups of concepts with similar
hierarchies. In many cases, additional information about the
relationships among the concepts is available through domain models,
such as Bayesian networks, decision trees, and association rules. A
domain model can be represented as a collection of {\em knowledge
rules}, each of which denotes a semantic relationship among several
concepts. These relationships may be complex, uncertain, and rely on
imprecise numeric values. In this paper, we introduce a new {\em
knowledge-based strategy} which uses the structure of these knowledge
rules as (soft) constraints on the alignment.

As a motivating example, consider two ontologies about basketball games.
One has datatype properties {\st height}, {\st weight} and a
binary property {\st center} for players, while the other has
the corresponding datatype properties {\st h}, {\st w}, and {\st
position}. Terminology-based strategies may not identify these
correspondences. However, if we know that a large value of {\st height}
implies {\st center} is true in the first ontology, and the same
relationship holds for {\st h} and {\st position} = {\st Center} in the
second ontology, then we tend to believe that {\st height} maps to {\st
h} and {\st center} maps to {\st position} = {\st Center}.

We use Markov logic networks (MLNs) \cite{Domingos09} as a probabilistic
language to combine the knowledge-based strategy with other strategies,
in a formalism similar to that of \cite{Niepert10}.  In particular, we
encode the knowledge-based strategy with weighted formulas that increase
the probability of alignments where corresponding concepts have
isomorphic relationships.  We use an MLN inference engine to find the
most likely alignment. We name our method Knowledge-Aware Ontology
Matching (KAOM).

% **************** DL: This paragraph mostly works, but it's a little
% strange.  First, it says our approach can handle complex
% correspondences.  Then, it says that our method has to be *extended*
% to handle complex correspondences (and this extended method is never
% named).  Furthermore, "simply extending our method to incorporate
% complex concepts" sounds like saying "we adapted our method to
% incorporate complex concepts," which is very generic -- you could
% almost say that about any modification to handle complex
% correspondences!  I understand that what you have in mind is much
% simpler, but the idea of how KAOM or a modification of it handles
% these cases doesn't really come through.  ****************
Our approach is also capable of identifying {\em complex
correspondences}, an extremely difficult task in ontology matching. A
complex correspondence is a correspondence between a simple concept and
a complex concept (e.g., {\st grad\_student} maps to the union of {\st
PhD} and {\st Masters}).  This can be achieved by constructing a set of
{\em complex concepts} (e.g., unions) in each ontology, subsequantly
generating candidate complex correspondences, and using multiple
strategies -- including the knowledge-based strategy -- to find the
correct complex correspondences.

The contributions of this work are as follows:

\begin{itemize}

\item We show how to represent common types of domain models as
knowledge rules, and how to use these knowledge rules to guide the
ontology matching process, leading to more accurate alignments. Our
approach is especially effective in identifying the correspondences of
numerical or nominal datatype properties.  By incorporating complex
concepts, our approach is also capable of discovering complex
correspondences, which is a very difficult scenario in the ontology
matching task.

\item We evaluate the effectiveness of KAOM in three domains with
different types of knowledge rules, and show that our approach not
only outperforms the state-of-the-art approaches for ontology matching
in one-to-one matching, but also discovers complex correspondences
successfully.

\end{itemize}

The rest of the paper is organized as follows. First we review pervious
work on ontology matching. We then introduce the concept of ``knowledge
rules'' with a definition and examples. Next, we show our approach of
using Markov logic to incorporate multiple strategies, including a
knowledge-based strategy and the treatment of complex correspondences.
Finally we present experimental results and conclude.

\section{Ontology Matching}
\label{sec:bg}

We begin by formally defining ontology matching.

\begin{definition}[Ontology Matching \cite{Euzenat07}]

Given two ontologies $O_1$ and $O_2$, a {\em correspondence} is a
3-tuple $\langle e_1, e_2, r \rangle$ where $e_1$ and $e_2$ are entities
of the first and second ontologies respectively, and $r$ is a semantic
relation such as equivalence ($\equiv$) and subsumptions ($\sqsubseteq$
or $\sqsupseteq$). An {\em alignment} is a set of correspondences. {\em
Ontology matching} is the task or process of identifying the correct
semantic alignment between the two ontologies. In most cases, ontology
matching focuses on equivalence relationships only.

\end{definition}

Most existing schema-level ontology matching systems use two types of
strategies: terminology-based and structure-based. Terminology-based
strategies are based on terminological similarity, such as string-based
or linguistic similarity measures. Structure-based strategies are based
on the assumption that two matching ontologies should have similar local
or global structures, where the structure is represented by subsumption
relationships of classes and properties, and domains and ranges of
properties. Advanced ontology matching systems often combine the two
types of strategies(e.g., \cite{Melnik02}). See \cite{Shvaiko11} for a
survey of ontology matching systems and algorithms. Recently, a
probabilistic framework based on Markov logic was proposed to combine
multiple strategies \cite{Niepert10}. 
%Cotterell13,Mao10,

%In particular, it encodes multiple strategies and heuristics into hard
%and soft constraints, and finds the best matching by minimizing the
%weighted number of violated constraints. The constraints include string
%similarity, the cardinality constraints which enforce that each concept
%matches at most one concept, the coherence constraints which prevent
%inconsistency induced by the matching, and the stability constraints
%which penalize dissimilar local subsumption relationships.

%  DL: Include a sentence like this, or not...?
% This framework will serve as the foundation of our ontology matching method.

\begin{definition}[Complex Correspondences]

A {\em complex concept} is a composition (e.g., unions, complements) of
one or more simple concepts. In
OWL\footnote{\url{http://www.w3.org/TR/owl2-primer/}}, there are several
constructors for creating complex classes and properties (see the top
part of Table \ref{tab:dl} for an incomplete list of constructors). A
{\em complex correspondence} is an equivalence relation between a simple
class or property and a complex class or property in two
ontologies \cite{Ritze08}.

\end{definition}

Previous work takes several different approaches to finding complex
correspondences (i.e., complex matching).  One is to construct
candidates for complex correspondences using operators for primitive
classes, such as string concatenation or arithmetic operations on
numbers \cite{Dhamankar04}.  
%An alternative pattern-based strategy is based on linguistic and
%structure features when a one-to-one alignment is given \cite{Ritze08}.  
\cite{Ritze08} introduce four specific patterns for complex
correspondences: Class by Attribute Type (CAT), Class by Inverse
Attribute Type (CIAT), Class by Attribute Value (CAV), and Property
Chain pattern (PC).  Finally, when aligned or overlapping data is
available, inductive logic programming (ILP) techniques can be used as
well \cite{Hu11,QinEtal07}.

%Many ontology matching systems make use of data instances to some extent
%(e.g., \cite{Dhamankar04,Doan02,Hu11,QinEtal07}). However, in this
%paper, we focus on the case where data are not available or data sharing
%is not preferred because of communication cost or privacy concerns. 

\section{Representation of Domain Knowledge}
\label{sec:rk}

In the AI community, knowledge is typically represented in {\em formal
languages}, among which ontologies and ontology-based (e.g., the Web
Ontology Language, OWL) languages are the most widely used forms. The
Web Ontology Language (OWL) is the W3C standard ontology language that
describes the classes, properties and relations of objects in a specific
domain. 

%In ontology languages such as OWL, knowledge is represented as logic
%{\em axioms}. These axioms describe properties of classes or relations
%(e.g., a relation is functional, symmetric, or antisymmetric, etc.), or
%a relationship of several entities (e.g., the relation `grandfather' is
%the composition of the two relations `father' and `parent').

OWL and many other ontology languages are based on variations of
description logics. However, the choice of using description logic as
the foundation of the Semantic Web ontology languages is largely due to
the trade-off between expressivity and reasoning efficiency. In tasks
such as ontology matching, reasoning does not need to be instant, so we
can afford to consider more general forms of knowledge outside of a
specific ontology language or description logic.

\begin{definition}[Knowledge Rule]

A knowledge rule is a sentence $R(a,b,\ldots; \theta)$ in a formal
language which consists of a relation $R$, a set of entities (i.e.,
classes, attributes or relations) $\{a,b,\ldots\}$, and (optionally) a
set of parameters $\theta$. A knowledge rule carries logical or
probabilistic semantics representing the relationship among these
entities.  The specific semantics depend on $R$.

\end{definition}

Many domain models and other types of knowledge can be represented as
sets of knowledge rules, each rule describing the relationship of a
small number of entities. The semantics of each relationship $R$ can
typically be expressed with a formal language. Table~\ref{tab:dl} shows
some examples of the symbols used in formal languages such as
description logic, along with their associated semantics.

\begin{table}[!htb]

\caption{Syntax and semantics of DL symbols (top), DL axioms (middle),
and other knowledge rules used in the examples of the paper
(bottom)}

\label{tab:dl}
\centering

\begin{tabular}{l | l}
\hline
Syntax & Semantics \\
\hline
%$\top$ & $\mathcal{D}$ \\
%$\bot$ & $\emptyset$ \\
$C \sqcap D$ & $C^\mathcal{I} \cap D^\mathcal{I}$ \\
$C \sqcup D$ & $C^\mathcal{I} \cup D^\mathcal{I}$ \\
$\neg C$ & $\mathcal{D} \backslash C^\mathcal{I}$ \\
%$\forall R.C$ & $\{x \in \mathcal{D} | \forall y ((x,y) \in R^\mathcal{I} \rightarrow y \in C^\mathcal{I}) \}$ \\
$\exists R.C$ & $\{x \in \mathcal{D} | \exists y ((x,y) \in R^\mathcal{I} \land y \in C^\mathcal{I}) \}$ \\
$R \circ S$ & $\{(x, y) | \exists z ((x,z) \in R^\mathcal{I} \land (z,y) \in S^\mathcal{I}) \}$ \\
%$R^-$ & $\{(x,y) | (y,x) \in R^\mathcal{I} \}$ \\
%$R \upharpoonleft C$ & $\{(x,y) \in R^\mathcal{I} | x \in C^\mathcal{I} \}$ \\
$R \downharpoonright C$ & $\{(x,y) \in R^\mathcal{I} | y \in C^\mathcal{I} \}$ \\
\hline
$C \sqsubseteq D$ & $C^\mathcal{I} \subseteq D^\mathcal{I}$ \\
$C \sqsubseteq \neg D$ & $C^\mathcal{I} \cap D^\mathcal{I} = \emptyset$ \\
\hline
$R \prec S$ & $y < y'$ for $\forall (x, y) \in R^\mathcal{I} \land (x, y') \in S^\mathcal{I}$ \\
$C \Rightarrow D$ & $\text{Pr}(D^\mathcal{I} | C^\mathcal{I})$ is close to 1 \\
\hline
\end{tabular}

\end{table}

We illustrate a few forms of knowledge rules with the following
examples. 
%For each rule, we provide a description in English, a
%logical representation, and an encoding as a knowledge rule with a
%particular semantic relationship, $R_i$.  We define a new relationship
%in each example, but, in a large domain model, most relationships
%would be appear many times in different rules.

\begin{example}
\label{ex:conf-deadline}
The submission deadline precedes the camera ready deadline:
$\textst{paperDueOn} \prec \textst{manuscriptDueOn}$
This is represented as $R_1(\textst{paperDueOn}, \textst{manuscriptDueOn})$
with $R_1(a, b): a \prec b$.
\end{example}

\begin{example}
\label{ex:nba-center}
A basketball player taller than 81 inches and heavier than
245 pounds is likely to be a center: 
$\textst{h} > 81 \wedge \textst{w} > 245 \Rightarrow \textst{pos} = \textst{Center}$
This rule can be viewed as a branch of a {\em decision tree} or an {\em
association rule}. It can be represented as $R_2(\textst{h}, \textst{w},
\textst{pos=Center},[81,245])$, with $R_2(a,b,c,\theta): a > \theta_1
\wedge b > \theta_2 \Rightarrow c$.

\end{example}

\begin{example}
\label{ex:social-smoke}
A smoker's friend is likely to be a smoker as well:
$
\textst{Smokes}(x) \wedge \textst{Friend}(x,y) \Rightarrow \textst{Smokes}(y)
$
%Relational rules such as this one describe relationships of attributes
%across multiple tables, as opposed to propositional data mining rules
%that are restricted to a single table. 
This rule can be represented as $R_3(\textst{Smoke}, \textst{Friend})$
with $R_3(a, b): a(x) \wedge b(x, y) \Rightarrow a(y)$.

\end{example}

%For the remainder of this paper, we will assume that the knowledge in
%both domains is represented as knowledge rules, as described in this
%section. 

\section{Knowledge Aware Ontology Matching}
\label{sec:mln}

In this section, we present our approach, Knowledge Aware Ontology
Matching (KAOM). KAOM uses Markov logic networks \cite{Domingos09} to
solve the ontology matching task. The MLN formulation is similar to
\cite{Niepert10} but incorporates the knowledge-based matching strategy
and treatment of complex correspondences.

%An MLN \cite{Domingos09} is a set of weighted formulas in first-order
%logic. Given a set of constants for individuals in a domain, an MLN
%induces a probability distribution over Herbrand interpretations or
%``possible worlds''.  
In the ontology matching problem, we represent a correspondence with a
binary relation, ${\tt match}(a,a')$, which is true if concept $a$
from the first ontology is semantically equivalent to concept $a'$
from the second ontology (e.g., ${\tt match}(\textst{writePaper}, \textst{writes})$ means
$\textst{writePaper} \equiv \textst{writes}$). 
%Each possible world therefore corresponds to an alignment of the two
%ontologies.  We want to find the most probable possible world, which is
%the configuration that maximizes the sum of weights of satisfied
%formulas.

We define three components of the MLN of the ontology matching problem:
{\em constants}, {\em evidence} and {\em formulas}. The logical
constants are the entities in both ontologies. The evidence includes the
complete set of OWL-supported relationships among all concepts in each
ontology (e.g., subsumptions and disjointness, we use an OWL reasoner to
create the complete set of OWL axioms.) , and rules converted to
first-order atoms as described in the previous section. 

For the formulas, we begin with a set of formulas adapted
from \cite{Niepert10}. The numbers preceding the formulas are the
weights. A missing weight means a formula with infinity weight.
\begin{enumerate}
\item \textbf{A-priori similarity}
is the string similarity between all pairs of concepts:
\begin{align}
s_{a,a'} \quad & {\tt match}(a, a')
\end{align}
where $s_{a,a'}$ is the string similarity between $a$ and $a'$,
which also serves as the weight of the formula.  We use
the Levenshtein measure \cite{Levenshtein66} for simple correspondences. This atomic
formula increases the probability of matching pairs of concepts with
similar strings, all other things being equal.

\item \textbf{Cardinality constraints}
enforce one-to-one simple (or complex) correspondences:
\begin{align}
{\tt match}(a, a') \land {\tt match}(a,a'') \Rightarrow a' = a''
\end{align}

\item \textbf{Coherence constraints}
enforce consistency of subclass relationships:
\begin{align}
a \sqsubseteq b \land a' \sqsubseteq \neg b' \Rightarrow \lnot ({\tt match}(a, a') \land {\tt match}(b,b'))
\end{align}

%\item \textbf{Stability constraints} enforce
%consistency of the subclass relationships between the two ontologies.
%They can be viewed as a special case of the knowledge-based
%constraints we introduce below.
\end{enumerate}

\subsection{Knowledge-based Strategy}

We propose a new {\em knowledge-based strategy} for ontology matching
that uses the similarity of knowledge rules in the two ontologies. It is
inspired by the structure-based strategy in many ontology matching
algorithms (e.g., \cite{Melnik02}). The strategy favors the alignments
that preserve the same types of knowledge rules, which extends the
subsumption relationship of entities in structure-based strategies 
%We now describe how we incorporate knowledge-based constraints into the
%MLN formulation through new formulas relating knowledge rules to
%matchings.  The {\em stability} constraints in \cite{Niepert10}
%consider three subclass relationships, including $a$ is a subclass of
%$b$ ({\st subclass}), and $a$ is a subclass or superclass of the domain
%or range of a property $b$ ({\st domainsub}, {\st rangesub}). We extend
%the relationships (knowledge rule patterns) 
to sub-property, disjoint properties, and user-defined relations such as
ordering of dates, and non-deterministic relationships such as
correlation and anti-correlation. The strategy can be represented as the
Markov logic formulas:
\begin{align}
-w_k & & & R_k(a,b,...) \land \neg R_k(a',b',...) \Rightarrow \nonumber \\
& & & {\tt match}(a,a') \land {\tt match}(b,b') \land ..., k = 1, ..., m \label{eq:mln2}
\end{align}
\begin{align}
+w_k' & & & R_k(a,b, ...) \land R_k(a',b', ...) \Rightarrow \nonumber \\
& & & {\tt match}(a,a') \land {\tt match}(b,b') \land ..., k = 1, ..., m \label{eq:mln3}
\end{align}
where $R_k$ is a rule pattern. 
% User-defined relations include those
%derived from decision trees, association rules, expert systems, and
%other knowledge sources outside the ontology.

%Besides the stability constraints, we introduce a new group of {\em
%similarity} constraints that encourage knowledge rules with the same
%pattern to have corresponding concepts.

\begin{example}
\label{ex:op1}

A reviewer of a paper cannot be the paper's author. In the {\st cmt}
ontology we have $R_4(\textst{writePaper}, \textst{readPaper})$ and in
the {\st confOf} ontology we have $R_4(\textst{write},
\textst{reviews})$ where $R_4(a,b): a \sqsubseteq \neg b$ is the
disjoint relationship of properties. Applying the constraint formulas
defined above, we increase the score of all alignments containing the
two correct correspondences: $\textst{writePaper} \equiv
\textst{writes}$ and $\textst{readPaper} \equiv \textst{reviews}$.

\end{example}

Rules involving continuous numerical attributes often include parameters
(e.g., thresholds in Example~\ref{ex:nba-center}) that do not match
between different ontologies. In order to apply the knowledge-based
strategy to numerical attributes, we make the assumption that
corresponding numerical attributes roughly have a {\em positive linear}
transformation. This assumption is often true in real applications, for
instance, when an imperial measure of height matches to a metric measure
of height. 

We propose two methods to handle numerical attributes.
The first method is to compute a {\em distance measure} (e.g.,
Kullback-Leibler divergence) between the distributions of the
corresponding attributes in a candidate alignment. 
Specifically, we replace Formulas~\ref{eq:mln2} and \ref{eq:mln3} with:
\begin{align}
d_0 - d & & & {\tt match}(a,a') \land {\tt match}(b,b') ..., k = 1, ..., m \label{eq:mln4}
\end{align}
where $d$ is a distance measure of the two rules $R_k(a,b,...)$ and
$R'_k(a',b',...)$ and $d_0$ is a threshold.

\begin{example}
\label{ex:nba-center2}
In the {\st nba-os} ontology, we have conditional rules converted from a decision tree, such as
\[
\textst{h} > 81 \wedge \textst{w} > 245 \Rightarrow \textst{Center}
\]
Similarly, in the {\st nbayahoo} ontology, we have 
\[
\textst{h'} > 2.06 \wedge \textst{w'} > 112.5 \Rightarrow \textst{Center'}
\]
Here the knowledge rules represent the conditional distributions of
multiple entities. 
We define the distance between the two conditional distributions as 
\begin{align*}
& d(\textst{h},\textst{w},\textst{Center};\textst{h'},\textst{w'},\textst{Center'})  \\
= & \mathbb{E}_{p(\textst{h},\textst{w})} d(p(\textst{Center} | \textst{h}, \textst{w}) || p(\textst{Center'} | \textst{h'}, \textst{w'}))
\end{align*}
where $\mathbb{E}(\cdot)$ is expectation and $d(p || p')$ is a distance
measure (Because $\textst{Center}$ and $\textst{Center}'$ are binary
attributes, we simply use $|p - p'|$ as the distance measure. For
numerical attributes, we can use the difference of two distribution
histograms as the distance measure). We assume the attribute
correspondences ($\textst{h}$ and $\textst{h'}$, $\textst{w}$ and
$\textst{w'}$) are linear mappings, and the linear relation can be
roughly estimated (e.g., by 
%using linear regression if $p(\textst{h})$ and $p(\textst{h'})$ are
%known, or simply 
matching the minimum and maximum numbers in these rules).  When
computing the expectation over $\textst{h}$ and $\textst{w}$, we apply
the linear mapping to generate corresponding values of $\textst{h'}$ and
$\textst{w'}$.
%, e.g., $\textst{h'} = 0.025 \textst{h}$, $\textst{w'} =
%0.45 \textst{w}$.  
The distribution of the conditional attributes
$p(\textst{h},\textst{w})$ can be roughly estimated as independent and
uniform over the ranges of the attributes.

\end{example}

The second method for handling continuous attributes is to discretize
%We handle the continuous attributes by discretizing
them, reducing the continuous attribute problem to the discrete
problem described earlier.  For example, suppose each continuous
attribute $x$ is replaced with a discrete attribute $x^d$, indicating
the quartile of $x$ rather than its original value. Then we have
$R_5(\textst{h}^d, \textst{w}^d, \textst{Center})$
and 
$R_5(\textst{h'}^d, \textst{w'}^d, \textst{Center'})$
with relation
$R_5(a, b, c): a=4 \land b=4 \Rightarrow c$,
and the discrete value of 4 indicates that both $a$ and $b$ are in the
top quartile. Other discretization methods are also possible, as long as the
discretization is done the same way (e.g., equal-width) in both domains.

Our method does not rely on the forms of knowledge rules, nor does it
rely on the algorithms used to learn these rules. As long as similar
data mining techniques or tools are used on both sides of ontologies, we
would always be able to find interesting knowledge-based similarities
between the two ontologies.

\subsection{Complex Correspondences}

Our approach can also find complex correspondences, which contain
complex concepts in either or both of the ontologies. We add the complex
concepts into consideration and treat them the same way as simple
concepts, and all the simple and complex correspondences will be solved
jointly by considering terminology, structure, and knowledge-based
strategies in a single probabilistic formulation. 

First, because complex concepts are recursively defined and potentially
infinite, we need to select a finite subset of complex concepts and use
them to generate the candidate correspondences. We will only include the
complex concepts occurring in the ontology axioms or in the knowledge
rules.

Second, we need to define a string similarity measure for each type of
complex correspondence. For example, \cite{Ritze08} requires two
conditions for a Class by Attribute Type (CAT) matching pattern $O_1:a
\equiv O_2: \exists p.b$ (e.g., $a$ = {\st Accepted\_Paper}, $p$ = {\st
hasDecision}, $b$ ={\st Acceptance}): $a$ and $b$ are terminologically
similar, and the domain of $p$ ({\st Paper} in the example) is a
superclass of $a$. We can therefore define the string similarity of $a$
and $\exists p.b$ to be the string similarity of $a$ and $b$ which
coincides with the first condition, and the second condition is encoded
in the structure stability constraints. The string similarity measure of
many other types of correspondences can be defined similarly based on
the heuristic method in \cite{Ritze08}. If there does not exist a
straight-forward way to define the string similarity for a certain type
of complex correspondences, we can simply set it to 0 and rely on other
strategies to identify such correspondences.

Lastly, we need constraints for the correspondence of two complex concepts. The
corresponding component concepts and same constructor always implies the
corresponding complex concepts, while in the other direction, it is a soft
constraint.
\begin{align*}
& \quad & {\tt match}(a, a') \land {\tt match}(b, b') \land ... \Rightarrow & {\tt match}(c, c') & \quad & \\
w^c_k & \quad & {\tt match}(a, a') \land {\tt match}(b, b') \land ... \Leftarrow & {\tt match}(c, c') & \quad &
\end{align*}
where $c = \text{cons}_k(a, b, ...), c' = \text{cons}_k(a', b', ...)$
for each constructor $\text{cons}_k$ (e.g., union, $\exists p.b$).

Some complex correspondences are almost impossible to be identified with
traditional strategies. With the knowledge-based strategy, it becomes
possible.

\begin{example}
\label{ex:pc1}
A reviewer of a paper cannot be the paper's author. In the {\st cmt} ontology we have
\[
\textst{writePaper} \sqsubseteq \neg \textst{readPaper}
\]
and in the {\st conference} ontology we have
\begin{align*}
\textst{contributes} \downharpoonright \textst{Reviewed\_contribution} \\
\sqsubseteq \neg (\textst{contributes} \circ \textst{reviews})
\end{align*}
We first build two complex concepts $\textst{contributes} \downharpoonright
\textst{Reviewed\_contribution}$ and $\textst{contributes} \circ
\textst{reviews}$. With $R_4(a, b) = a \sqsubseteq \neg b$ (disjoint
properties), the score function would favor the correspondences
\begin{align*}
\textst{writePaper} & \equiv \textst{contributes} \downharpoonright \textst{Reviewed\_contribution} \\
\textst{readPaper} & \equiv \textst{contributes} \circ \textst{reviews}
\end{align*}

\end{example}

%To handle complex correspondences, we add constants to the domain for
%constructions that occur in knowledge rules, and add knowledge rules
%that contain these new complex concepts.  We enforce type constraints
%between the simple and complex concepts, as described in the previous
%section.  We use the previously described methods for string
%similarity between complex and simple concepts.  For complex to
%complex correspondences, the string similarity measure is zero, but we
%have constraints

\section{Experiments}
\label{sec:exp}

We test our KAOM approach on three domains: NBA, census and conference. The sizes
of the ontologies of these domains are listed in Table~\ref{tab:sizes}. These
domains contain very different forms of ontologies and knowledge rules, so we
can examine the generality and robustness of our approach.

\begin{table}
\centering

\caption{Number of classes (\#c), object properties (\#o), data properties (\#d) and
nominal values (\#v) of each ontology used in the experiments.}

\label{tab:sizes}
\begin{tabular}{c | c | c | c | c | c}
\hline
domain & ontology & \#c & \#o & \#d & \#v \\
\hline
\multirow{2}{*}{NBA} & nba-os & 3 & 3 & 20 & 3 \\
& yahoo & 4 & 4 & 21 & 7 \\
\hline
\multirow{2}{*}{census} & adult & 1 & 0 & 15 & 101 \\
& income & 1 & 0 & 12 & 97 \\
\hline
\multirow{5}{*}{OntoFarm} & cmt & 36 & 50 & 10 & 0 \\
& confOf & 38 & 13 & 25 & 0 \\
& conference & 60 & 46 & 18 & 6 \\
& edas & 103 & 30 & 20 & 0 \\
& ekaw & 78 & 33 & 0 & 0 \\
\hline
\end{tabular}
\end{table}

We use Pellet \cite{Sirin07} for logical inference of the ontological
axioms and
TheBeast\footnote{\url{http://code.google.com/p/thebeast/}} \cite{Riedel08}
and Rockit\footnote{\url{https://code.google.com/p/rockit/}. We use
RockIt for the census domain because TheBeast is not able to handle the
large number of rules in that domain.} \cite{Noessner13} for Markov
logic inference. 
%We ran all experiments on a machine with 24 Intel Xeon E5-2640 cores
%@2500 MHz and 8GB memory. 
We compare our system (KAOM) with three others: KAOM without the
knowledge-based strategy (MLOM), CODI \cite{Huber11} (a new version of
\cite{Niepert10}, which is essentially a different implementation of
MLOM), and logmap2 \cite{Jimenez-Ruiz11}, a top performing system in
OAEI 2014~\footnote{\url{http://oaei.ontologymatching.org/2014/}}. 

We manually specify the weights of the Markov logic formlas in KAOM and
MLOM. The weights of stability constraints for subclass relationships
are set with values same as the ones used in \cite{Niepert10}, i.e., the
weight for subclass is -0.5, and those for sub-domain and range are
-0.25. In KAOM, we also set the weights for different types of
similarity rules based on our assessment of their relative importance
and kept these weights fixed during the experiments.

\subsection{NBA}

The NBA domain is a simple setting that we use to demonstrate the
effectiveness of our approach. We collected data from the NBA official
website and the Yahoo NBA website. 
%These two datasets contain information about players, teams, matches,
%arenas, etc.
For each ontology, we used the WinMine
toolkit\footnote{\url{http://research.microsoft.com/en-us/um/people/dmax/WinMine/Tooldoc.htm}}
to learn a decision tree for each attribute using the
other attributes as inputs.

%DD: we may need to give a reference for probabilistic decision tree.
% SJ: It is just decision tree.

For each pair of conditional distributions based on decision tree with up
to three attributes, we calculate their similarity based on the distance measure described in
% DL: I checked Example 5, and it doesn't completely explain histogram
% intersection distance.  I don't think many readers will be able to
% guess exactly what you mean here.  Clarify (either here or near
% Example 5 somewhere).
Example~\ref{ex:nba-center2}. We use the Markov logic formula
(\ref{eq:mln4}) with the threshold $d_0 = 0.2$. To make the task more
challenging, we did not use any name similarity measures. Our method
% DL: "most" --> define most!  5 out of 7?  12 out of 13?  80%?  60%?
% A quantitative answer here will be stronger.
successfully identified the correspondence of all the numerical and nominal
attributes, including height, weight and positions (center, forward and
guard) of players. In contrast, without a name similarity measure, no
other method can solve the matching problem at all.

\subsection{Census}

We consider two census datasets and their ontologies from UC Irvine data
repository\footnote{\url{https://archive.ics.uci.edu/ml/datasets.html}}. Both
datasets represent census data but are sampled and post-processed differently.
These two census ontologies are flat with a single concept but many datatype
properties and nominal values. For this domain, we use association rules as
the knowledge.
%DD: we may need to give a reference for association rule mining.
% SJ: Association rule is already mentioned in previous sections. Does it need a reference
% here?
We first discretize each numerical attribute into five
intervals, and then generate association rules for each ontology using the
Apriori algorithm with a minimum confidence of 0.9 and minimum support of
0.001. For example, one generated rule is:
\begin{alltt}\small
age='(-inf-25.5]' education='11th' 
   hours-per-week= '(-inf-35.5]'
=> gross-income='<=50K' conf:(1)
\end{alltt}
This is represented as
\[
R_6(\textst{age}^d, \textst{11th}, \textst{hours-per-week}^d, \textst{gross-income}^d)
\]
where $x^d$ refers to the discretized value of $x$, split into one
fifth percentile intervals, and $R_6(a, b, c, d): a=1 \land b \land c=1 \Rightarrow d=1$.
%Two rules match only when the corresponding numerical values are in the same
%percentile interval and the nominal values have corresponding values. 
For scalability reasons, we consider up to three concepts in a knowledge rule,
i.e., association rules with up to three attributes. We set the weight of
knowledge similarity constraints for the association rules to 0.25.

\begin{figure}[!htb]
%\begin{minipage}[t]{0.32\linewidth}
\centering
\includegraphics[width=0.8\columnwidth]{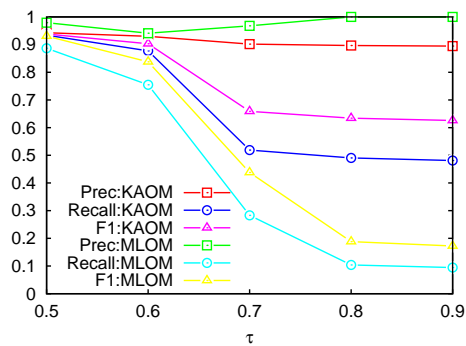}
\caption{Precision, recall and F1 on the census domain as a function of the string similarity
threshold $\tau$.}
\label{fig:census}
%\end{minipage}
\end{figure}

\begin{figure*}
\begin{minipage}[t]{0.32\linewidth}
\includegraphics[width=\columnwidth]{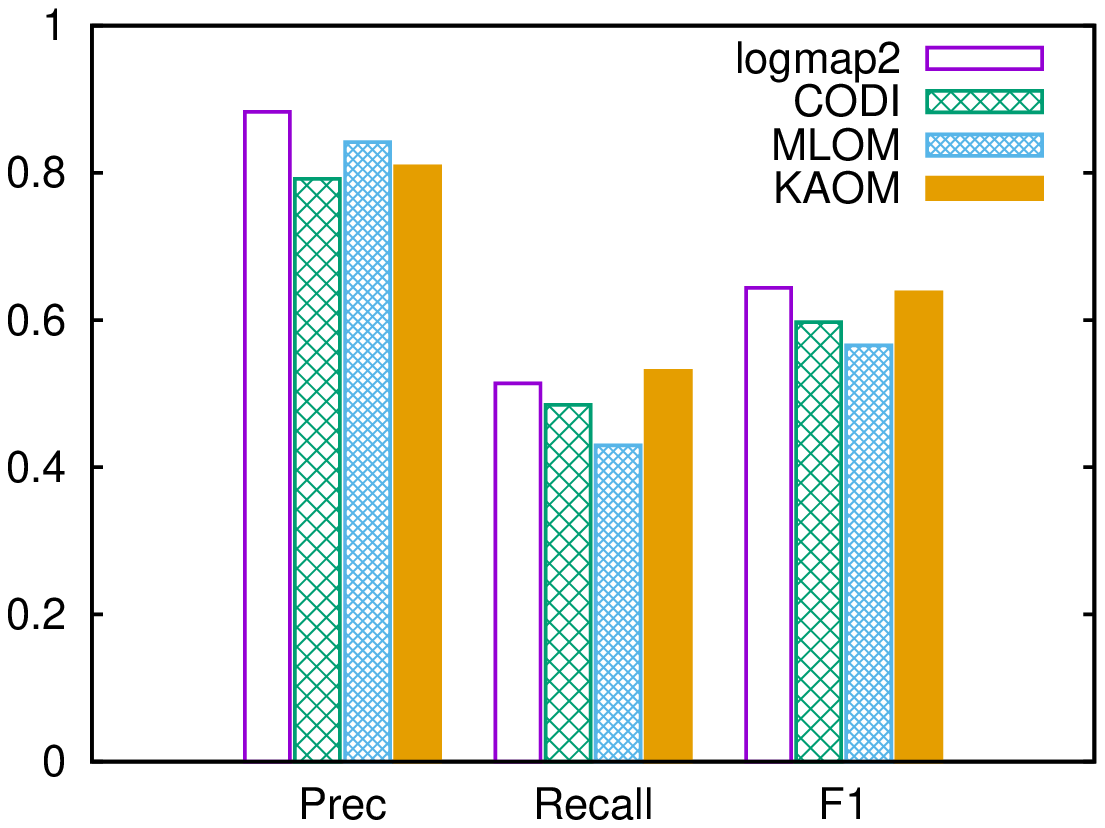}
\caption{Precision, recall and F1 on the OntoFarm domain with only the one-to-one correspondences.}
\label{fig:ontofarm}
\end{minipage}
\begin{minipage}[t]{0.32\linewidth}
\centering
\includegraphics[width=\columnwidth]{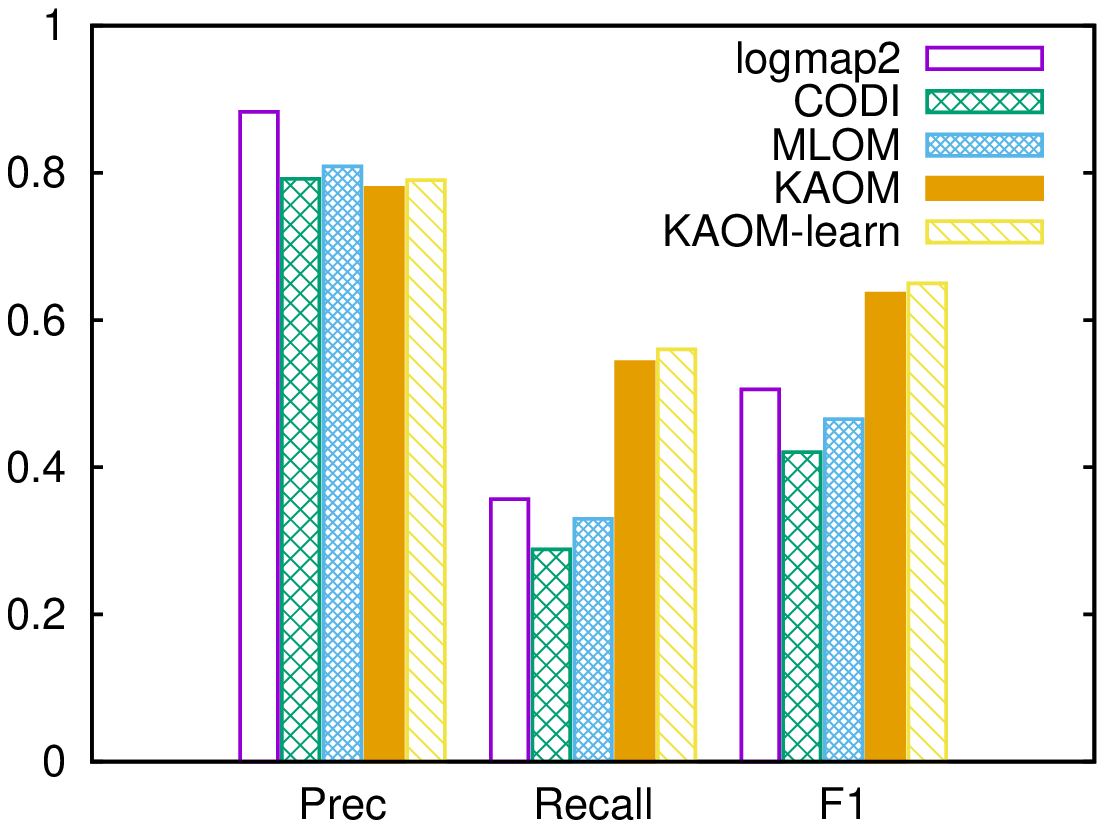}
\caption{Precision, recall and F1 on the OntoFarm domain with the complex correspondences.}
\label{fig:ontofarm-cc}
\end{minipage}
\begin{minipage}[t]{0.32\linewidth}
\centering
\includegraphics[width=\columnwidth]{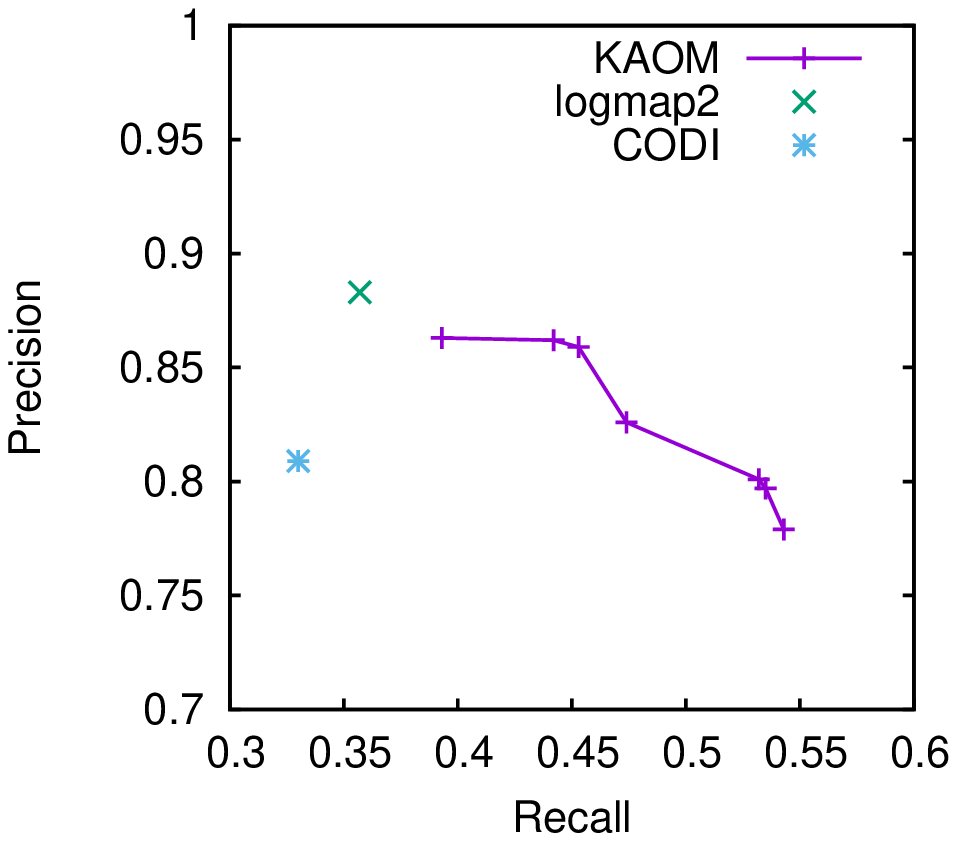}
\caption{Precision-recall curve on the OntoFarm domain with the complex correspondences.}
\label{fig:ontofarm-cc-pr}
\end{minipage}
\end{figure*}

In the Markov logic formulation in \cite{Niepert10}, only the correspondences
with apriori similarity measure larger than a threshold $\tau$ are added as
evidence. 
%This parameter is used to trade off between scalability and accuracy.  If
%$\tau$ is set too low, there will be many initial apriori constraints so the
%inference runs more slowly. If $\tau$ is set too high, we may miss some
%correct correspondences since they are not even candidates. When the knowledge
%similarity constraints are considered, we essentially added more
%correspondences as candidates based on their knowledge-based similarity.
In the experiments, we set $\tau$ with different values from 0.50 to 0.90. When
$\tau$ is large, we deliberately discard the string similarity information for
some correspondences. MLOM for this task is an extension of \cite{Niepert10} by
adding correspondences of {\em nominal values} and their dependencies with the
related attributes. The results are shown in Figure~\ref{fig:census}. We can
see that KAOM always gets better recall and F1, with only a slight degradation
in precision. This means our approach fully leverages the knowledge rule
information and thus does not rely too much on the names of the concepts to
determine the mapping. For example, when $\tau$ is 0.70, KAOM finds
% DL: "a majority" --> Why not be specific and list actual numbers???
6 out of 8 correspondences of values of {\st adult:workclass} and {\st
income:class\_of\_worker}, while MLOM finds none. The other two systems were not
designed for nominal value correspondences, but CODI only identified 7
% DL: How many attribute correspondences is "all"?
and logmap2 only identified 3 attribute correspondences, while KAOM and MLOM find all the 12
attribute correspondences.

\subsection{OntoFarm}

In order to show how our system can use manually created expert
knowledge bases, we use OntoFarm, a standard ontology matching benchmark
for an academic conference domain as the third domain in our experiments. As part of OAEI,
it has been widely used in the evaluation of ontology matching systems.
The process of manually knowledge rule creation is time consuming, so we
only used 5 of the OntoFarm ontologies ({\st cmt}, {\st conference},
{\st confOf}, {\st edas}, {\st ekaw}).  Using their knowledge of
computer science conferences and the structure of just one ontology, two
individuals listed a number of rules (e.g.,
Example~\ref{ex:conf-deadline}). We then translated these rules into
each of the five ontologies. Thus, the same knowledge was added to each
of the ontologies, but its representation depended on the specific
ontology.  For some ontologies, some of the rules were not representable
with the concepts in them and thus had to be omitted. This manually
constructed knowledge base was developed before running any experiments
and kept fixed throughout our experiments. Among the 5 ontologies, we
have 10 pairs of matching tasks in total. We set $\tau$ to 0.70, and the
weight for the knowledge similarity constraints to 1.0.

We first compare the four methods to the reference one-to-one alignment
from the benchmark (Figure~\ref{fig:ontofarm}). KAOM achieves similar
precision and F1, and better recall than other systems. It was able to
identify correspondences in which the concept names are very different,
for instance, {\st cmt:readPaper} $\equiv$ {\st confOf:reviews}. Note
that the similarity constraints work in concert with other constraints.
For instance, in Example~\ref{ex:op1}, since disjointness is a symmetric
knowledge rule, domain and range constraints could be helpful to
identify whether {\st cmt:writePaper} should match to {\st
confOf:writes} or {\st confOf:reviews}.

To evaluate our approach on complex correspondences, we extended the
reference alignment with hand-labeled complex correspondences
(Figure~\ref{fig:ontofarm-cc}). MLOM does not perform well in this task
because the complex correspondences require a good similarity measure to
become candidates (such as the linguistic features in \cite{Ritze08}).
KAOM, however, uses the structure of the rules to find many complex
correspondences without relying on complex similarity measures.
%(Note that the numbers in the one-to-one and complex correspondence
%tasks are not directly comparable, since the gold-standard alignments
%are different.)
For this task we also tried learning the weights of the formulas~\footnote{We used
MIRA implemented in TheBeast for weight learning.}
(KAOM-learn). For each of the 10 pairs of ontologies, we used the rest 9
pairs as training data. KAOM-learn performs slightly better than KAOM.
% DL: What algorithm and implementation did you use for MLN weight learning?
% DL: Also, the discussion of learning should probably go back when we
% describe our approach...

With the hand-picked or automatically learned weights, KAOM produces a
single most-likely alignment.  However, we can further tune KAOM to
produce alignments with higher recall or higher precision.  We
accomplish this by adding the MLN formula ${\tt match}(a, a')$ with
weight $w$.  When $w$ is positive, alignments with more matches are
more likely, and when $w$ is negative, alignments with fewer matches
are more likely (all other things being equal).  We adjusted this
weight to produce the precision-recall curve shown in
Figure~\ref{fig:ontofarm-cc-pr}.  KAOM dominates CODI and provides
much higher recall values than logmap2, although logmap2's best
precision remains slightly above KAOM's.
\section{Conclusion}
\label{sec:concl}

% DL: I think this could be expanded / revised / improved some.  ...but
% it's somewhat tricky to write, and I can't figure it out right now.
We proposed a new ontology matching algorithm KAOM. The key component of
KAOM is the knowledge-based strategy, which is based on the intuition
that ontologies about the same domain should contain similar knowledge
rules, in spite of the different terminologies they use. KAOM is also
capable of discovering complex correspondences, by treating complex
concepts the same way as simple ones. We encode the knowledge-based
strategy and other strategies in Markov logic and find the best
alignment with its inference tools. Experiments on the datasets and
ontologies from three different domains show that our method effectively
uses knowledge rules of different forms to outperform several
state-of-the-art ontology matching methods.

\paragraph{Acknowledgement} This research is funded by NSF grant IIS-1118050.

\bibliographystyle{aaai}
\bibliography{kaom}

\begin{thebibliography}{}

\bibitem[\protect\citeauthoryear{Dhamankar \bgroup et al\mbox.\egroup
  }{2004}]{Dhamankar04}
Dhamankar, R.; Lee, Y.; Doan, A.; Halevy, A.; and Domingos, P.
\newblock 2004.
\newblock {iMAP}: Discovering complex semantic matches between database
  schemas.
\newblock In {\em Proceedings of the 2004 ACM SIGMOD International Conference
  on Management of Data}, SIGMOD '04,  383--394.

\bibitem[\protect\citeauthoryear{Domingos and Lowd}{2009}]{Domingos09}
Domingos, P., and Lowd, D.
\newblock 2009.
\newblock {\em {M}arkov Logic: An Interface Layer for Artificial Intelligence}.
\newblock Synthesis Lectures on Artificial Intelligence and Machine Learning.
  Morgan \& Claypool.

\bibitem[\protect\citeauthoryear{Euzenat and Shvaiko}{2007}]{Euzenat07}
Euzenat, J., and Shvaiko, P.
\newblock 2007.
\newblock {\em Ontology Matching}.
\newblock Secaucus, NJ, USA: Springer-Verlag New York, Inc.

\bibitem[\protect\citeauthoryear{Hu \bgroup et al\mbox.\egroup }{2011}]{Hu11}
Hu, W.; Chen, J.; Zhang, H.; and Qu, Y.
\newblock 2011.
\newblock Learning complex mappings between ontologies.
\newblock In {\em Proceedings of JIST},  350--357.

\bibitem[\protect\citeauthoryear{Huber \bgroup et al\mbox.\egroup
  }{2011}]{Huber11}
Huber, J.; Sztyler, T.; Noessner, J.; and Meilicke, C.
\newblock 2011.
\newblock {CODI}: Combinatorial optimization for data integration--results for
  {OAEI} 2011.
\newblock {\em Ontology Matching}  134.

\bibitem[\protect\citeauthoryear{Jim{\'e}nez-Ruiz, Grau, and
  Zhou}{2012}]{Jimenez-Ruiz11}
Jim{\'e}nez-Ruiz, E.; Grau, B.~C.; and Zhou, Y.
\newblock 2012.
\newblock {LogMap} 2.0: Towards logic-based, scalable and interactive ontology
  matching.
\newblock In {\em Proceedings of the 4th International Workshop on Semantic Web
  Applications and Tools for the Life Sciences}, SWAT4LS '11,  45--46.

\bibitem[\protect\citeauthoryear{Levenshtein}{1966}]{Levenshtein66}
Levenshtein, V.
\newblock 1966.
\newblock Binary codes capable of correcting deletions, insertions and
  reversals.
\newblock {\em Soviet Physics Doklady} 10:707.

\bibitem[\protect\citeauthoryear{Melnik, Garcia-Molina, and
  Rahm}{2002}]{Melnik02}
Melnik, S.; Garcia-Molina, H.; and Rahm, E.
\newblock 2002.
\newblock Similarity flooding: A versatile graph matching algorithm.
\newblock In {\em Proceedings of Eighteenth International Conference on Data
  Engineering}.

\bibitem[\protect\citeauthoryear{Niepert, Meilicke, and
  Stuckenschmidt}{2010}]{Niepert10}
Niepert, M.; Meilicke, C.; and Stuckenschmidt, H.
\newblock 2010.
\newblock A probabilistic-logical framework for ontology matching.
\newblock In Fox, M., and Poole, D., eds., {\em Proceedings of the 24th AAAI
  Conference on Artificial Intelligence},  1413--1418.

\bibitem[\protect\citeauthoryear{Noessner, Niepert, and
  Stuckenschmidt}{2013}]{Noessner13}
Noessner, J.; Niepert, M.; and Stuckenschmidt, H.
\newblock 2013.
\newblock {RockIt}: Exploiting parallelism and symmetry for {MAP} inference in
  statistical relational models.
\newblock In {\em Proceedings of the Twenty-Seventh {AAAI} Conference on
  Artificial Intelligence}.

\bibitem[\protect\citeauthoryear{Qin, Dou, and LePendu}{2007}]{QinEtal07}
Qin, H.; Dou, D.; and LePendu, P.
\newblock 2007.
\newblock Discovering executable semantic mappings between ontologies.
\newblock In {\em Proceedings of ODBASE},  832--849.

\bibitem[\protect\citeauthoryear{Riedel}{2008}]{Riedel08}
Riedel, S.
\newblock 2008.
\newblock Improving the accuracy and efficiency of {MAP} inference for {M}arkov
  logic.
\newblock In {\em Proceedings of the Proceedings of the 24th Conference on
  Uncertainty in Artificial Intelligence (UAI-08)},  468--475.

\bibitem[\protect\citeauthoryear{Ritze \bgroup et al\mbox.\egroup
  }{2008}]{Ritze08}
Ritze, D.; Meilicke, C.; Sváb-Zamazal, O.; and Stuckenschmidt, H.
\newblock 2008.
\newblock A pattern-based ontology matching approach for detecting complex
  correspondences.
\newblock In Shvaiko, P.; Euzenat, J.; Giunchiglia, F.; Stuckenschmidt, H.;
  Noy, N.~F.; and Rosenthal, A., eds., {\em OM}, volume 551 of {\em CEUR
  Workshop Proceedings}.

\bibitem[\protect\citeauthoryear{Shvaiko and Euzenat}{2011}]{Shvaiko11}
Shvaiko, P., and Euzenat, J.
\newblock 2011.
\newblock Ontology matching: State of the art and future challenges.
\newblock {\em IEEE Transactions on Knowledge and Data Engineering} PP(99).

\bibitem[\protect\citeauthoryear{Sirin \bgroup et al\mbox.\egroup
  }{2007}]{Sirin07}
Sirin, E.; Parsia, B.; Grau, B.~C.; Kalyanpur, A.; and Katz, Y.
\newblock 2007.
\newblock Pellet: A practical {OWL-DL} reasoner.
\newblock {\em Web Semant.} 5(2):51--53.

\end{thebibliography}

\end{document}